\documentclass[runningheads]{llncs}
\usepackage{graphicx}
\usepackage{booktabs}
\usepackage{multirow}
\usepackage{stmaryrd}
\usepackage[colorlinks,linkcolor=blue]{hyperref}
\usepackage{amsmath,amssymb} 
\usepackage{color}

\def\ie{{\it i.e.}}

\def\et{{\it et al.~}}
\def\etc{{\it etc.}}

\def\I{{\mathbf I}}
\def\A{{\mathbf A}}
\def\W{{\mathbf W}}
\def\P{{\mathbf P}}
\def\V{{\mathbf V}}
\def\X{{\mathbf X}}
\def\Y{{\mathbf Y}}
\def\Z{{\mathbf Z}}

\begin{document}
\pagestyle{headings}
\mainmatter
\def\ECCV20SubNumber{1543}  

\title{Edge-aware Graph Representation Learning and Reasoning for Face Parsing} 




\author{Gusi Te\inst{1} \and
Yinglu Liu\inst{2} \and
Wei Hu\inst{1}\thanks{Corresponding author: Wei Hu (forhuwei@pku.edu.cn). This work was in collaboration with JD AI Research during Gusi Te's internship there. This work was supported by National Natural Science Foundation of China [61972009], Beijing Natural Science Foundation [4194080] and Beijing Academy of Artificial Intelligence (BAAI). } \and
Hailin Shi\inst{2} \and
Tao Mei\inst{2}}

%
%
\institute{Wangxuan Institute of Computer Technology, Peking University, Beijing, China \\
\email{\{tegusi, forhuwei\}@pku.edu.cn} \and
JD AI Research, Beijing, China\\
\email{\{liuyinglu1,shihailin,tmei\}@jd.com}}

\maketitle

\begin{abstract}
Face parsing infers a pixel-wise label to each facial component, which has drawn much attention recently. 
Previous methods have shown their efficiency in face parsing, which however overlook the correlation among different face regions. 
The correlation is a critical clue about the facial appearance, pose, expression, \textit{etc.}, and should be taken into account for face parsing.
To this end, we propose to model and reason the region-wise relations by learning graph representations, and leverage the edge information between regions for optimized abstraction.     
Specifically, we encode a facial image onto a global graph representation where a collection of pixels (``regions") with similar features are projected to each vertex.
Our model learns and reasons over relations between the regions by propagating information across vertices on the graph. 
Furthermore, we incorporate the edge information to aggregate the pixel-wise features onto vertices, which emphasizes on the features around edges for fine segmentation along edges. 
The finally learned graph representation is projected back to pixel grids for parsing.  
Experiments demonstrate that our model outperforms state-of-the-art methods on the widely used Helen dataset, and also exhibits the superior performance on the large-scale CelebAMask-HQ and LaPa dataset. The code is available at \href{https://github.com/tegusi/EAGRNet}{https://github.com/tegusi/EAGRNet}.

\end{abstract}

\section{Introduction}

Face parsing assigns a pixel-wise label to each semantic component, such as facial skin, eyes, mouth and nose, which is a particular task in semantic segmentation. 
It has been applied in a variety of scenarios such as face understanding, editing, synthesis, and animation \cite{CelebAMask-HQ,zhang2018synthesis,zhang2016joint}.  

The region-based methods have been recently proposed to model the facial components separately \cite{lin2019face,yin2020end,liu2017face}, and achieved state-of-the-art performance on the current benchmarks. 
However, these methods are based on the individual information within each region, and the correlation among regions is not exploited yet to capture long range dependencies. 
In fact, facial components present themselves with abundant correlation between each other. 
For instance, eyes, mouth and eyebrows will generally become more curvy when people smile; facial skin and other components will be dark when the lighting is weak, and so on.



The correlation between the facial components is the critical clue in face representation, and should be taken into account in the face parsing.
To this end, we propose to learn graph representations over facial images, which model the relations between regions and enable reasoning over non-local regions to capture long range dependencies. 
To bridge the facial image pixels and graph vertices, we project a collection of pixels (a ``region") with similar features to each vertex. 
The pixel-wise features in a region are aggregated to the feature of the corresponding vertex. 
In particular, to achieve accurate segmentation along the edges between different components,  we propose the edge attention in the pixel-to-vertex projection, assigning larger weights to the features of edge pixels during the feature aggregation. 
Further, the graph representation learns the relations between facial regions, \ie, the graph connectivity between vertices, and reasons over the relations by propagating information across all vertices on the graph, which is able to capture long range correlations in the facial image. 
The learned graph representation is finally projected back to the pixel grids for face parsing. 
Since the number of vertices is significantly smaller than that of pixels, the graph representation also reduces redundancy in features as well as computational complexity effectively.

Specifically, given an input facial image, we first encode the high-level and low-level feature maps by the ResNet backbone \cite{he2016deep}. 
Then, we build a projection matrix to map a cluster of pixels with similar features to each vertex. 
The feature of each vertex is taken as the weighted aggregation of pixel-wise features in the cluster, where features of edge pixels are assigned with larger weights via an edge mask. 
Next, we learn and reason over the relations between vertices (\ie, regions) via graph convolution \cite{henaff2015deep,defferrard2016convolutional} to further extract global semantic features. 
The learned features are finally projected back to a pixel-wise feature map. 
We test our model on Helen, CelebAMask-HQ and LaPa datasets, and surpass state-of-the-art methods.

Our main contributions are summarized as follows.
\begin{itemize}
    \item We propose to exploit the relations between regions for face parsing by modeling on a region-level graph representation, where we project a collection of pixels with similar features to each vertex and reason over the relations to capture long range dependencies.
    
    \item We introduce edge attention in the pixel-to-vertex feature projection, which emphasizes on features of edge pixels during the feature aggregation to each vertex and thus enforces accurate segmentation along edges. 
    \item We conduct extensive experiments on Helen, CelebAMask-HQ and LaPa datasets. The experimental results show our model outperforms state-of-the-art methods on almost every category.
\end{itemize}

\section{Related Work}


\subsection{Face Parsing}
Face parsing is a division of semantic segmentation, which assigns different labels to the corresponding regions on human faces, such as nose, eyes, mouth and \etc. The methods of face parsing could be classified into global-based and local-based methods. 

Traditionally, hand crafted features including SIFT \cite{smith2013exemplar} are applied to model the facial structure. Warrell \et describe spatial relationship of facial parts with epitome model \cite{warrell2009labelfaces}. Kae \et combine Conditional Random Field (CRF) with a Restricted Boltzmann Machine (RBM) to extract local and global features \cite{kae2013augmenting}. With the rapid development of machine learning, CNN has been introduced to learn more robust and rich features. Liu \et import CNN-based features into the CRF framework to model individual pixel labels and neighborhood dependencies \cite{liu2015multi}. Luo \et propose a hierarchical deep neural network to extract multi-scale facial features \cite{luo2012hierarchical}. Zhou \et adopt adversarial learning approach to train the network and capture high-order inconsistency \cite{zhou2013extensive}. Liu \et design a CNN-RNN hybrid model that benefits from both high quality features of CNN and non-local properties of RNN \cite{liu2017face}. Zhou \et present an interlinked CNN that takes multi-scale images as input and allows bidirectional information passing \cite{zhou2015interlinked}. Lin \et propose a novel RoI Tanh-Warping operator preserving central and peripheral information. It contains two branches with the local-based for inner facial components and the global based for outer facial ones. This method shows high performance especially on hair segmentation \cite{lin2019face}.

\subsection{Attention Mechanism}
Attention mechanism has been proposed to capture long-range information \cite{bahdanau2014neural}, and applied to many applications such as sentence encoding \cite{vaswani2017attention} and image feature extraction \cite{wang2018non}. Limited by the locality of convolution operators, CNN lacks the ability to model global contextual information. Furthermore, Chen \et propose Double Attention Model that gathers information spatially and temporally to improve complexity of traditional non-local modules \cite{chen20182}. Zhao \et propose a point-wise spatial attention module, relaxing the local neighborhood constraint \cite{zhao2018psanet}. Zhu \et also present an asymmetric module to reduce abundant computation and distillate features \cite{zhu2019asymmetric}. Fu \et devise a dual attention module that applies both spatial and channel attention in feature maps \cite{fu2019dual}. To research underlying relationship between different regions, Chen \et project original features into interactive space and utilize GCN to exploit high order relationship \cite{chen2019graph}. Li \et devise a robust attention module that incorporates the Expectation-Maximization algorithm \cite{li2019expectation}.

\subsection{Graph Reasoning}
Interpreting images from the graph perspective is an interesting idea, since an image could be regarded as regular pixel grids. Chandra \et propose Conditional Random Field (CRF) based method on image segmentation \cite{chandra2017dense}. Besides, graph convolution network (GCN) is imported into image segmentation. Li \et introduce graph convolution to the semantic segmentation, which projects features into vertices in the graph domain and applies graph convolution afterwards \cite{li2018beyond}. Furthermore, Lu \et propose Graph-FCN where semantic segmentation is reduced to vertex classification by directly transforming an image into regular grids \cite{lu2019graph}. Pourian \et propose a method of semi-supervised segmentation \cite{pourian2015weakly}. The image is divided into community graph and different labels are assigned to corresponding communities. 
Te \et propose a computation-efficient and posture-invariant face representation with only a few key points on hypergraphs for face anti-spoofing beyond 2D attacks \cite{te2020exploring}.  
Zhang \et utilize graph convolution both in the coordinate space and feature space \cite{zhang2019dual}.

\section{Methods}

\begin{figure}[t]
    \centering
    \includegraphics[width=0.9\textwidth]{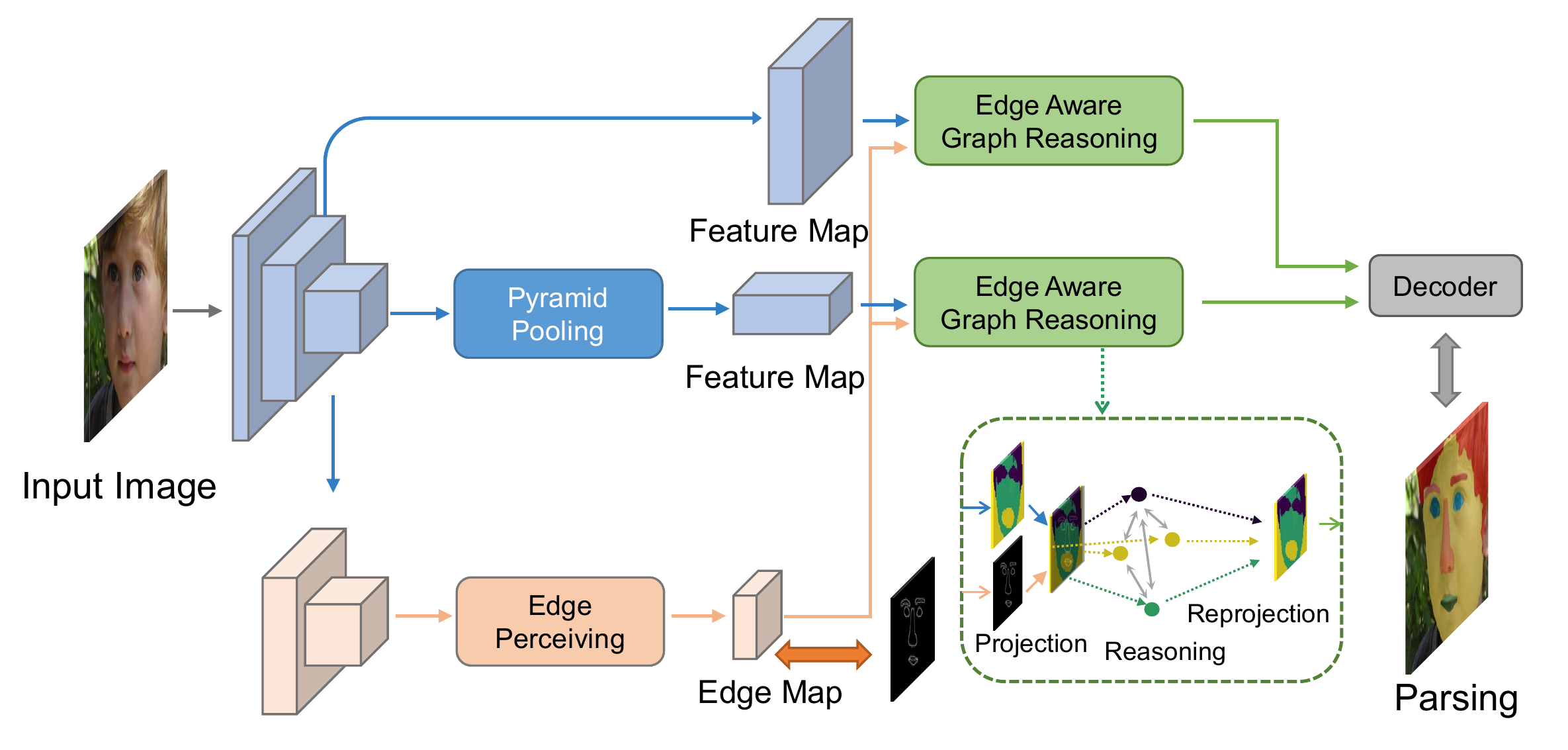}
    \caption{The overview of the proposed face parsing framework.}
    \label{fig:framework}
\end{figure}

\subsection{Overview}\label{sec3.1}

As illustrated in Fig.~\ref{fig:framework}, given an input facial image, we aim to predict the corresponding parsing label and auxiliary edge map. The overall framework of our method consists of three procedures as follows. 

\begin{itemize}
    \item \textbf{Feature and Edge Extraction.} We take ResNet as the backbone to extract features at various levels for multi-scale representation.  
    The low-level features contain more details but lack semantic information, while the high-level features provide rich semantics with global information at the cost of image details. To fully exploit the global information in high-level features, we employ a spatial pyramid pooling operation to learn multi-scale contextual information.
    Further, we construct an edge perceiving module to acquire an edge map for the subsequent module. 
    
    \item \textbf{Edge Aware Graph Reasoning.} We feed the feature map and edge map into the proposed Edge Aware Graph Reasoning (EAGR) module, aiming to learn intrinsic graph representations for the characterization of the relations between regions. 
    The EAGR module consists of three operations: graph projection, graph reasoning and graph reprojection, which projects the original features onto vertices in an edge-aware fashion, reasons the relations between vertices (regions) over the graph and projects the learned graph representation back to pixel grids, leading to a refined feature map with the same size.
    
    \item \textbf{Semantic Decoding.} We fuse the refined features into a decoder to predict the final result of face parsing. The high-level feature map is upsampled to the same dimension as the low-level one. 
    We concatenate both feature maps and leverage 1 $\times$ 1 convolution layer to reduce feature channels, predicting the final parsing labels.
\end{itemize}



\subsection{Edge-Aware Graph Reasoning}\label{sec3.2}

Inspired by the non-local module \cite{wang2018non}, we aim to build the long-range interactions between distant regions, which is critical for the description of the facial structure. 
In particular, we propose edge-aware graph reasoning to model the long-range relations between regions on a graph, which consists of edge-aware graph projection, graph reasoning and graph reprojection.  

\subsubsection{Edge-Aware Graph Projection}
\label{subsubsec:projection}

We first revisit the typical non-local modules. 
Given a feature map $\X \in \mathbb{R}^{HW \times C}$,  where $H$ and $W$ refer to the height and width of the input image respectively and $C$ is the number of feature channels. 
A typical non-local module is formulated as:
\begin{equation}
\label{eq:non_local}
\widetilde{\X} = \text{softmax}\left(\theta(\X)\varphi^{\top}(\X)\right)\gamma(\X) = \V\gamma(\X),
\end{equation}
where $\theta$, $\varphi$ and $\gamma$ are convolution operations with $ 1 \times 1$ kernel size. 
$\V \in \mathbb{R}^{HW \times HW}$ is regarded as the attention maps to model the long-range dependencies. 
However, the complexity of computing $\V$ is $\mathcal{O}(H^2W^2C)$, which does not scale well with increasing number of pixels $HW$. 
To address this issue, we propose a simple yet effective edge-aware projection operation to eliminate the redundancy in features.

\begin{figure}[t]
    \centering
    \includegraphics[width=0.7\textwidth]{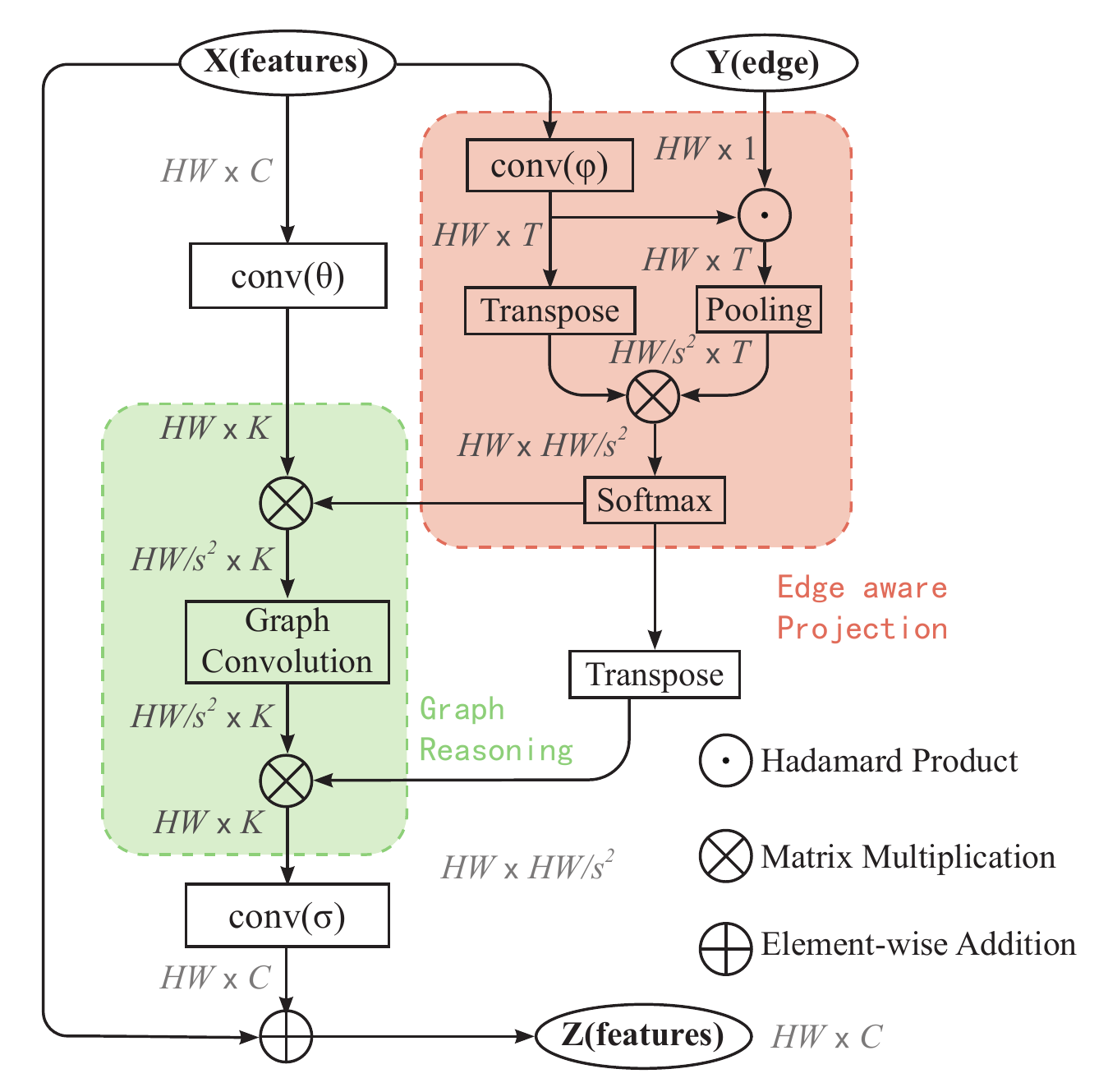}
    \caption{Architecture of the Edge Aware Graph Reasoning module.}
        \label{fig:module}
\end{figure}

Given an input feature map $\X \in \mathbb{R}^{HW \times C}$ and an edge map $\Y \in \mathbb{R}^{HW \times 1}$, we construct a projection matrix $\P$ by mapping $\X$ onto vertices of a graph with $\Y$ as a prior. 
Specifically, we first reduce the dimension of $\X$ in the feature space via a convolution operation $\varphi$ with $1 \times 1$ kernel size, leading to $\varphi(\X) \in \mathbb{R}^{HW \times T} $, $T < C$. 
Then, we duplicate the edge map $Y$ to the same dimension of $\varphi(\X)$ for ease of computation. 
We incorporate the edge information into the projection, by taking the Hadamard Product of $\varphi(\X)$ and $\Y$. 
As the edge map $\Y$ encodes the probability of each pixel being an edge pixel, the Hadamard Product operation essentially assigns a weight to the feature of each pixel, with larger weights to features of edge pixels. 
Further,  we introduce an average {\it pooling} operation $\mathcal P(\cdot)$ with stride $s$ to obtain anchors of vertices. 
These anchors represent the centers of each region of pixels, and we take the multiplication of $\varphi(\X)$ and anchors to capture the similarity between anchors and each pixel. 
We then apply a softmax function for normalization. 
Formally, the projection matrix takes the form:
\begin{equation}
 \P = \text{softmax}\left(\mathcal{P}(\varphi(\X) \odot \Y) \cdot \varphi(\X)^{\top}\right),
 \label{eq:projection_matrix}
\end{equation}
where $\odot$ denotes the Hadamard product, and $\P \in \mathbb{R}^{HW/s^2 \times HW}$. 

In Eq.~\eqref{eq:projection_matrix}, we have two critical operations: the edge attention and the pooling operation. 
The edge attention emphasizes the features of edge pixels by assigning larger weights to edge pixels.
Further, we propose the pooling operation in the features, whose benefits are in twofold aspects. 
On one hand, the pooling leads to compact representations by averaging over features to remove the redundancy. 
On the other hand, by pooling with stride $s$, the computation complexity is reduced from $\mathcal{O}(H^2W^2C)$ in non-local modules to $\mathcal{O}(H^2W^2C/s^2)$.

With the acquired projection matrix $\P$, we project the pixel-wise features $\X$ onto the graph domain, \ie, 
\begin{equation}
    \X_G = \P\theta(\X),
    \label{eq:projection_operation}
\end{equation}
where $\theta$ is a a convolution operation with $ 1 \times 1$ kernel size so as to reduce the dimension of $\X$, resulting in $\theta(\X) \in \mathbb{R}^{HW \times K}$.
The projection aggregates pixels with similar features as each anchor to one vertex, thus each vertex essentially represents a region in the facial images. 
Hence, we bridge the connection between pixels and each region via the proposed edge aware graph projection, leading to the features of the projected vertices on the graph $\X_G \in \mathbb{R}^{HW/s^2 \times K}$ via Eq.~\eqref{eq:projection_operation}. 

\subsubsection{Graph Reasoning}

Next, we learn the connectivity between vertices from $\X_G$, \ie, the relations between regions. 
Meanwhile, we reason over the relations by propagating information across vertices to learn higher-level semantic information. 
This is elegantly realized by a single-layer Graph Convolution Network (GCN). 
Specifically, we feed the input vertex features $\X_G$ into a first-order approximation of spectral graph convolution. 
The output feature map $\hat{\X}_G \in \mathbb{R}^{HW/s^2 \times K}$ is
\begin{equation}
    \hat{\X}_G = \text{ReLU}\left[(\I - \A)\X_G\W_G\right] = \text{ReLU}\left[(\I - \A)\P\theta(\X)\W_G\right],
\end{equation}
where $\A$ denotes the adjacent matrix that encodes the graph connectivity to learn, $\W_G \in \mathbb{R}^{K \times K}$ denotes the weights of the GCN, and ReLU is the activation function. 
The features $\hat{\X}_G $ are acquired by the vertex-wise interaction (multiplication with $(\I - \A)$) and channel-wise interaction (multiplication with $\W_G$). 

Different from the original one-layer GCN \cite{kipf2016semi} in which the graph $\A$ is hand-crafted, we randomly initialize $\A$ and learn from vertex features.
Moreover, we add a residual connection to reserve features of raw vertices. 
Based on the learned graph, the information propagation across all vertices leads to the finally reasoned relations between regions.  
After graph reasoning, pixels embedded within one vertex share the same context of features modeled by graph convolution. 
We set the same number of output channels as the input to keep consistency, allowing the module to be compatible with the subsequent process.

\subsubsection{Graph Reprojection}

In order to fit into existing framework, we reproject the extracted vertex features in the graph domain to the original pixel grids.
Given the learned graph representation $\hat{\X}_G \in \mathbb{R}^{HW/s^2 \times K}$, we aim to compute a matrix $\mathbf{V} \in \mathbb{R}^{HW \times HW/s^2}$ that maps $\hat{\X}_G$ to the pixel space.
In theory, $\mathbf{V}$ could be taken as the inverse of the projection matrix $\P$. However, it is nontrivial to compute because $\P$ is not a square matrix. 
To tackle this problem, we take the transpose matrix $\P^{\top}$ as the reprojection matrix \cite{li2018beyond}, in which $\P^{\top}_{ij}$ reflects the correlation between vertex $i$ and pixel $j$. The limitation of this operation is that the row vectors in $\P^{\top}$ are not normalized.

After reprojection, we deploy a $1 \times 1$ convolution operation $\sigma$ to increase the feature channels in consistent with the input features $\X$. 
Then, we take the summation of the reprojected refined features and the original feature map as the final features.
The final pixel-wise feature map $\Z \in \mathbb{R}^{HW \times C}$ is thus computed by 
\begin{equation}
    \Z = \X + \sigma(\P^{\top}\hat{\X}_G).
\end{equation}

\subsection{The Loss Function}
To further strengthen the effect of the proposed edge aware graph reasoning, we introduce the boundary-attention loss (BA-Loss) inspired by \cite{liu2020new} besides the traditional cross entropy loss for predicted parsing maps and edge maps.
The BA-loss computes the loss between the predicted label and the ground truth only at edge pixels, thus improving the segmentation accuracy of critical edge pixels that are difficult to distinguish.
Mathematically, the BA-loss is written as 
\begin{equation}
    \mathcal{L}_{\text{BA}} = \sum_{i=1}^{HW}\sum_{j=1}^{N}\left[ e_i = 1 \right] y_{ij}\log{p_{ij}},
\end{equation}
where $i$ is the index of pixels, $j$ is the index of classes and $N$ is the number of classes. $e_i$ denotes the edge label, $y_{ij}$ denotes the ground truth label of face parsing, and $p_{ij}$ denotes the predicted parsing label. 
$\left[ \cdot \right]$ is the Iverson bracket, which denotes a number that is $1$ if the condition in the bracket is satisfied, and $0$ otherwise. 

The total loss function is then defined as follows:
\begin{equation}
    \mathcal{L} = \mathcal{L}_{\text{parsing}} + \lambda_1\mathcal{L}_{\text{edge}} + \lambda_2\mathcal{L}_{\text{BA}},
\end{equation}
where $\mathcal{L}_{\text{parsing}}$ and $\mathcal{L}_{\text{edge}}$ are classical cross entropy losses for the parsing and edge maps.
$\lambda_1$ and $\lambda_2$ are two hyper-parameters to strike a balance among the three loss functions.  

\subsection{Analysis} \label{sec3.3}
Since non-local modules and graph-based methods have drawn increasing attention, it is interesting to analyze the similarities and differences between previous works and our method. 

\subsubsection{Comparison with non-local modules}
Typically, a traditional non-local module models {\it pixel-wise} correlations by feature similarities.  
However, the high-order relationship between regions are not captured. 
In contrast, we exploit the correlation among distinct regions via the proposed graph projection and reasoning. 
The features of each vertex embed not only local contextual anchor aggregated by average pooling in a certain region but also global features from the overall pixels. 
We further learn and reason over the relations between regions by graph convolution, which captures high-order semantic relations between different facial regions.  

Also, the computation complexity of non-local modules is expensive in general as discussed in Section~\ref{subsubsec:projection}.  
Our proposed edge-aware pooling addresses the issue by extracting significant anchors to replace redundant query points. 
Also, we do not incorporate pixels within each facial region during the sampling process while focusing on edge pixels, thus improving boundary details. 
The intuition is that pixels within each region tend to share similar features.


\subsubsection{Comparison with graph-based models}
In comparison with other graph-based models, such as \cite{chen2019graph,li2018beyond}, we improve the graph projection process by introducing locality in sampling in particular. 
In previous works, each vertex is simply represented by a weighted sum of image pixels, which does not consider edge information explicitly and brings ambiguity in understanding vertices. 
Besides, with different inputs of feature maps, the pixel-wise features often vary greatly but the projection matrix is fixed after training. 
In contrast, we incorporate the edge information into the projection process to emphasize on edge pixels, which preserves boundary details well. 
Further, we specify vertex anchors locally based on the average pooling, which conforms with the rule that the location of facial components keeps almost unchanged after face alignment. 

\section{Experiments}

\subsection{Datasets and Metrics}

The Helen dataset includes 2,330 images with 11 categories: background, skin, left/right brow, left/right eye, upper/lower lip, inner mouth and hair. Specifically, we keep the same train/validation/test protocol as in \cite{le2012interactive}. The number of the training, validation and test samples are 2,000, 230 and 100, respectively.
The CelebAMask-HQ dataset is a large-scale face parsing dataset which consists of 24,183 training images, 2,993 validation images and 2,824 test images.
The number of categories in CelebAMask-HQ is 19. In addition to facial components, the accessories such as eyeglass, earring, necklace, neck, and cloth are also annotated in the CelebAMask-HQ dataset.
The LaPa dataset is a newly released challenging dataset for face parsing, which contains 11 categories as Helen, covering large variations in facial expression, pose and occlusion. It consists of 18,176 training images, 2,000 validation images and 2,000 test images.

During training, we use the rotation and scale augmentation. The rotation angle is randomly selected from $(-30^\circ, 30^\circ)$ and the scale factor is randomly selected from $(0.75, 1.25)$.
The edge mask is extracted according to the semantic label map. If the label of a pixel is different with its 4 neighborhoods, it is regarded as a edge pixel.
For the Helen dataset, similar to \cite{lin2019face}, we implement face alignment as a pre-processing step and the results are re-mapped to the original image for evaluation.

We employ three evaluation metrics to measure the performance of our model: pixel accuracy, mean intersection over union (mIoU) and F1 score.
Directly employing the accuracy metric ignores the scale variance amid facial components, while the mean IoU and F1 score are better for evaluation. 
To keep consistent with the previous methods, we report the overall F1-score on the Helen dataset, which is computed over the merged facial components: brows (left+right), eyes (left+right), nose, mouth (upper lip+lower lip+inner mouth). For the CelebAMask-HQ and LaPa datasets, the mean F1-score over all categories excluding background is employed.
\subsection{Implementation Details}
Our backbone is a modified version of the ResNet-101 \cite{he2016deep} excluding the average pooling layer, and the Conv1 block is changed to three $3 \times 3$ convolutional layers. 
For the pyramid pooling module, we follow the implementation in \cite{zhao2017pyramid} to exploit global contextual information. The pooling factors are $\{1, 2, 3, 6\}$. 
Similar to \cite{ruan2019devil}, the edge perceiving module predicts a two-channel edge map based on the outputs of Conv2, Conv3 and Conv4 in ResNet-101. 
The outputs of Conv1 and the pyramid pooling serve as the low-level and high-level feature maps, respectively. Both of them are fed into the EAGR module separately for graph representation learning. 

As for the EAGR module, we set the pooling size to $6 \times 6$. To pay more attention on the facial components, we just utilize the central $4 \times 4$ anchors for graph construction. The feature dimensions $K$ and $T$ are set to 128 and 64, respectively.

Stochastic Gradient Descent (SGD) is employed for optimization. We initialize the network with a pretrained model on ImageNet. The input size is $473 \times 473$ and the batch size is set to 28. The learning rate starts at 0.001 with the weight decay of 0.0005. 
The batch normalization is implemented with In-Place Activated Batch Norm \cite{rotabulo2017place}.

\subsection{Ablation study}

\begin{table}[t]
\centering

\caption{Ablation study on the Helen dataset.}
\begin{tabular}{c|ccccc|ccc}
\toprule
Model & Baseline & Edge & Graph & Reasoning & BA-loss & mIoU & F1-score &  Accuracy \\ 
\midrule
1 & \checkmark &  &   &  &  & 76.5 & 91.4 & 85.9 \\ 
2 & \checkmark & \checkmark & &  &   & 77.5 & 92.0 & 86.2 \\ 
3 & \checkmark &  & \checkmark & \checkmark &  & 77.3 & 92.3 & 85.8 \\ 
4 & \checkmark & \checkmark & \checkmark & \checkmark &  & 77.8 & 92.4 & 84.6 \\ 
5 & \checkmark & \checkmark & \checkmark & & \checkmark & 77.3 & 92.3 & 86.7 \\ 
6 & \checkmark & \checkmark & \checkmark & \checkmark & \checkmark & 78.2 & 92.8 & 87.3 \\ 
\bottomrule

\end{tabular}
\label{table:ablation}
\end{table}

\begin{figure}[t]
    \centering
    \includegraphics[width=\textwidth]{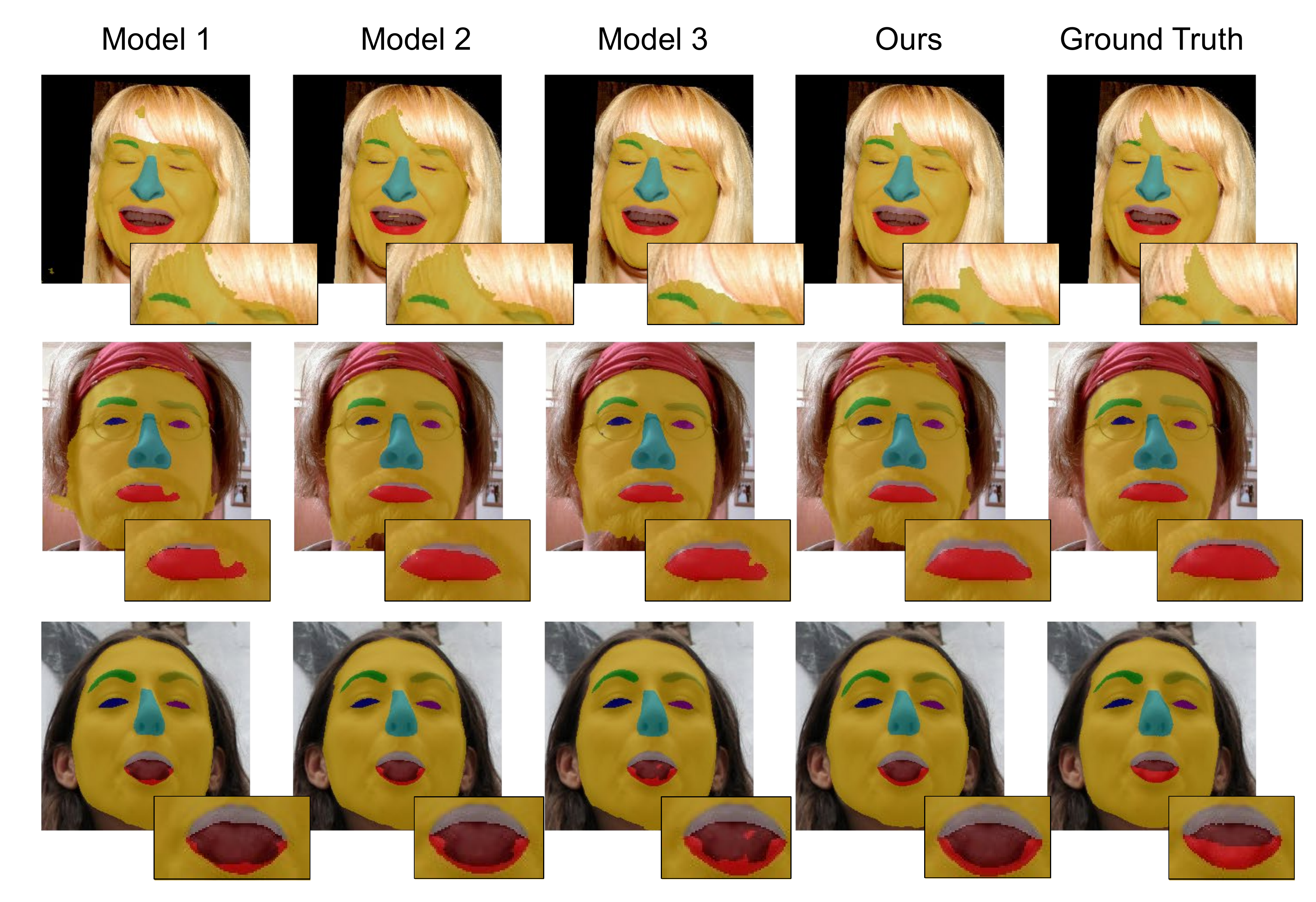}
    \caption{Parsing results of different models on the Helen dataset. (Best viewed in color)}
    \label{fig:ablation_result}
\end{figure}

\subsubsection{On different components}

We demonstrate the effectiveness of different components in the proposed EAGR module. 
Specifically, we remove some components and train the model from scratch under the same initialization. 
The quantitative results are reported in Table~\ref{table:ablation}. {\it Baseline} means the model only utilizes the ResNet backbone, pyramid pooling and multi-scale decoder without any EGAR module, and {\it Edge} represents whether edge aware pooling is employed.
{\it Graph} represents the EAGR module, while {\it Reasoning} indicates the graph reasoning excluding graph projection and reprojection. 
We observe that {\it Edge} and {\it Graph} lead to improvement over the baseline by $1\%$ in mIoU respectively. 
When both components are taken into account, we achieve even better performance. 
The boundary-attention loss (BA-loss) also leads to performance improvement. 

We also provide subjective results of face parsing from different models in Fig.~\ref{fig:ablation_result}.
Results of incomplete models exhibit varying degrees of deficiency around edges in particular, such as the edge between the hair and skin in the first row, the upper lips in the second row, and edges around the mouse in the third row. 
In contrast, our complete model produce the best results with accurate edges between face constitutes, which is almost the same as the ground truth. 
This validates the effectiveness of the proposed edge aware graph reasoning. 

\begin{table}[t]
\centering
\caption{Performance comparison with different deployment of the EAGR module and pooling size.}
\label{table:modules}
\begin{tabular}{c|c|c|c|c|c|c|c}
\toprule
\begin{minipage}{2cm} \centering  \end{minipage} & \multicolumn{4}{c|}{Deployment} & \multicolumn{3}{c}{Pooling Size} \\
\hline
Model & ~0-module~ & ~1-module~ & ~2-modules~ & ~3-modules~ &  ~4 $\times$ 4~ & 6 ~$\times$ 6~ & ~8 $\times$ 8~ \\ 
\hline
mIoU      & 77.6 & 77.6 &  78.2 & 77.4 &  77.0 & 78.2 & 78.0 \\ 
F1-score    & 92.0 & 92.5 & 92.8 & 92.3 &  92.1 & 92.8 & 92.6 \\ 
Accuracy & 85.5 & 86.0 & 87.3 & 85.4 &  87.4 & 87.3 & 87.0\\ 
\bottomrule

\end{tabular}
\end{table}

\subsubsection{On the deployment of the EAGR module}

We also conduct experiments on the deployment of the EAGR module with respect to the feature maps as well as pooling sizes. 
We take the output of Conv2 in the ResNet as the low-level feature map, and that of the pyramid pooling module as the high-level feature map.  
We compare four deployment schemes: 1) $0$-module, where no EAGR module is applied; 
2) $1$-module, where the low-level and high-level feature maps are concatenated, and then fed into an EAGR module;
3) $2$-modules, where the low-level and high-level feature maps are fed into one EAGR module respectively;
4) $3$-moduels, which combines 2) and 3).   
As listed in Table~\ref{table:modules}, the scheme of $2$-modules leads to the best performance, which is the one we finally adopt. 

We also test the influence of the pooling size, where the number of vertices changes along with the pooling size. 
As presented in Table~\ref{table:modules}, the size of 6 $\times$ 6 leads to the best performance, while enlarging the pooling size further does not bring performance improvement. 
This is because more detailed anchors lead to the loss of integrity, which breaks the holistic semantic representation.

\subsubsection{On the complexity in time and space}
\begin{figure}[t]
    \centering
    \includegraphics[width=0.85\textwidth]{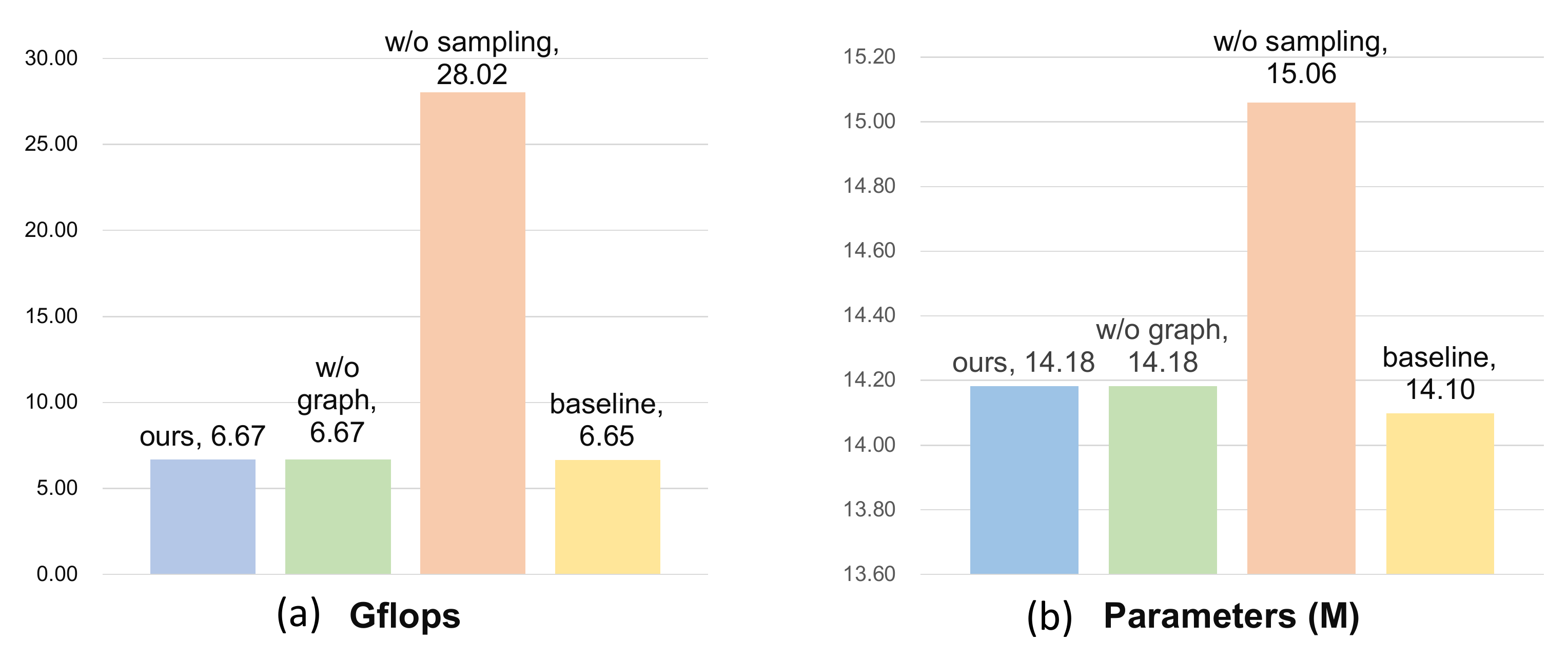}
    \caption{Complexity comparison on the Helen dataset. We reset the start value of y-axis for better appearance.}
    \label{fig:table}
\end{figure}

Further, we study the complexity of different models in time and space in Fig.~\ref{fig:table}. We compare with three schemes: 
1) a simplified version without the EAGR module, which we refer to as the \textit{Baseline}; 
2) a non-local module \cite{wang2018non} employed without edge aware sampling (\ie, pooling) as \textit{Without sampling}; and 
3) a version without graph convolution for reasoning as \textit{Without graph}. 
As presented in Fig.~\ref{fig:table}, compared with the typical non-local module, our proposed method reduces the computation time by more than 4$\times$ in terms of flops. 
We also see that the computation and space complexity of our method is comparable to those of the \textit{Baseline}, which indicates that most complexity comes from the backbone network.  Using Nvidia P40, the time cost of our model for a single image is $89$ms in the inference stage. 
This demonstrates that the proposed EAGR module achieves significant performance improvement with trivial computational overhead.

\subsection{Comparison with the state-of-the-art}

\begin{table}[t]
\centering
\caption{Comparison with state-of-the-art methods on the Helen dataset (in F1 score).}
\label{table:Helen}
\begin{tabular}{c|cccccccc|c}
\toprule
Methods & Skin & Nose & U-lip & I-mouth & L-lip & Eyes & Brows & Mouth & Overall \\ 
\midrule

Liu \et \cite{liu2017face} & 92.1 & 93.0 & 74.3 & 79.2 & 81.7 & 86.8 & 77.0 & 89.1 & 88.6 \\ 
Lin \et \cite{lin2019face} & 94.5 & 95.6 & 79.6 & 86.7 & 89.8 & 89.6 & 83.1 & 95.0 & 92.4 \\ 
Wei \et \cite{wei2019accurate} & \textbf{95.6} & 95.2 & 80.0 & 86.7 & 86.4 & 89.0 & 82.6 & 93.6 & 91.7 \\ 
Yin \et \cite{yin2020end} & - & \textbf{96.3} & 82.4 & 85.6 & 86.6 & 89.5 & 84.8 & 92.8 & 91.0 \\ 
Liu \et \cite{liu2020new} & 94.9 & 95.8 & \textbf{83.7} & 89.1 & \textbf{91.4} & 89.8 & 83.5 & \textbf{96.1} & 93.1 \\
\midrule
Ours & 94.6 & 96.1 & 83.6 & \textbf{89.8} & 91.0 & \textbf{90.2} & \textbf{84.9} & 95.5 & \textbf{93.2} \\
\bottomrule
\end{tabular}
\end{table}

\begin{table}[t]
\caption{Experimental comparison on the CelebAMask-HQ dataset (in F1 score).}
\label{table:Celeb}
\begin{tabular}{c|ccccccccc|c}
\toprule
\multirow{2}{*}{Methods}  & Face    & Nose  & Glasses & L-Eye & R-Eye & L-Brow  & R-Brow   & L-Ear & R-Ear & \multirow{2}{*}{Mean} \\
                          & I-Mouth & U-Lip & L-Lip   & Hair  & Hat   & Earring & Necklace & Neck  & Cloth &                       \\
\midrule
\multirow{2}{*}{Zhao \et \cite{zhao2017pyramid}}   & 94.8    & 90.3  & 75.8    & 79.9  & 80.1  & 77.3    & 78       & 75.6  & 73.1  & \multirow{2}{*}{76.2} \\
                          & 89.8    & 87.1  & 88.8    & 90.4  & 58.2  & 65.7    & 19.4     & 82.7  & 64.2  &                       \\
\midrule
\multirow{2}{*}{Lee \et\cite{CelebAMask-HQ}} & 95.5    & 85.6  & \textbf{92.9}    & 84.3  & 85.2  & 81.4    & 81.2     & 84.9  & 83.1  & \multirow{2}{*}{80.3} \\
                          & 63.4    & 88.9  & 90.1    & 86.6  & \textbf{91.3}  & 63.2    & 26.1     & \textbf{92.8}  & 68.3  &                       \\
\midrule
\multirow{2}{*}{Ours}     & \textbf{96.2}    & \textbf{94}    & 92.3    & \textbf{88.6}  & \textbf{88.7}  & \textbf{85.7}    & \textbf{85.2}     & \textbf{88}    & \textbf{85.7}  & \multirow{2}{*}{\textbf{85.1}} \\
                          & \textbf{95}      & \textbf{88.9}  & \textbf{91.2}    & \textbf{94.9}  & 87.6  & \textbf{68.3}    & \textbf{27.6}     & 89.4  & \textbf{85.3}  &    \\  
                          
\bottomrule
\end{tabular}
\end{table}

\begin{table}[t]
\centering
\caption{Experimental comparison on the LaPa dataset (in F1 score).}
\label{table:LaPa}
\begin{tabular}{c|cccccccccc|c}
\toprule
Methods & Skin & Hair & L-Eye & R-Eye & U-lip & I-mouth & L-lip & Nose & L-Brow & R-Brow & Mean \\ 
\midrule
Zhao \et \cite{zhao2017pyramid} & 93.5 & 94.1 & 86.3 & 86.0 & 83.6 & 86.9 & 84.7 & 94.8 & 86.8 & 86.9  & 88.4 \\
Liu \et \cite{liu2020new} & 97.2 & \textbf{96.3} & 88.1 & 88.0 &  84.4 & 87.6 & 85.7 & 95.5 & \textbf{87.7} & \textbf{87.6}  & 89.8 \\
\midrule
Ours & \textbf{97.3} & 96.2 & \textbf{89.5} & \textbf{90.0} & \textbf{88.1} & \textbf{90.0} & \textbf{89.0} & \textbf{97.1} & 86.5 & 87.0 & \textbf{91.1} \\
\bottomrule
\end{tabular}
\end{table}

We conduct experiments on the broadly acknowledged Helen dataset to demonstrate the superiority of the proposed model. 
To keep consistent with the previous works\cite{liu2017face,lin2019face,wei2019accurate,yin2020end,liu2020new}, we employ the overall F1 score to measure the performance, which is computed by combining the merged eyes, brows, nose and mouth categories. As Table~\ref{table:Helen} shows, Our model surpasses state-of-the-art methods and achieves $93.2\%$ on this dataset. 

We also evaluate our model on the newly proposed CelebAMask-HQ~\cite{CelebAMask-HQ} and LaPa~\cite{liu2020new} datasets, whose scales are about 10 times larger than the Helen dataset. Different from the Helen dataset, CelebAMask-HQ and LaPa have accurate annotation for hair. Therefore, mean F1-score (over all foreground categories) is employed for better evaluation. Table~\ref{table:Celeb} and Table~\ref{table:LaPa} give the comparison results of the related works and our method on these two datasets, respectively.

\subsection{Visualization of Graph Projection}

\begin{figure}[t]
    \centering
    \includegraphics[width=0.7\textwidth]{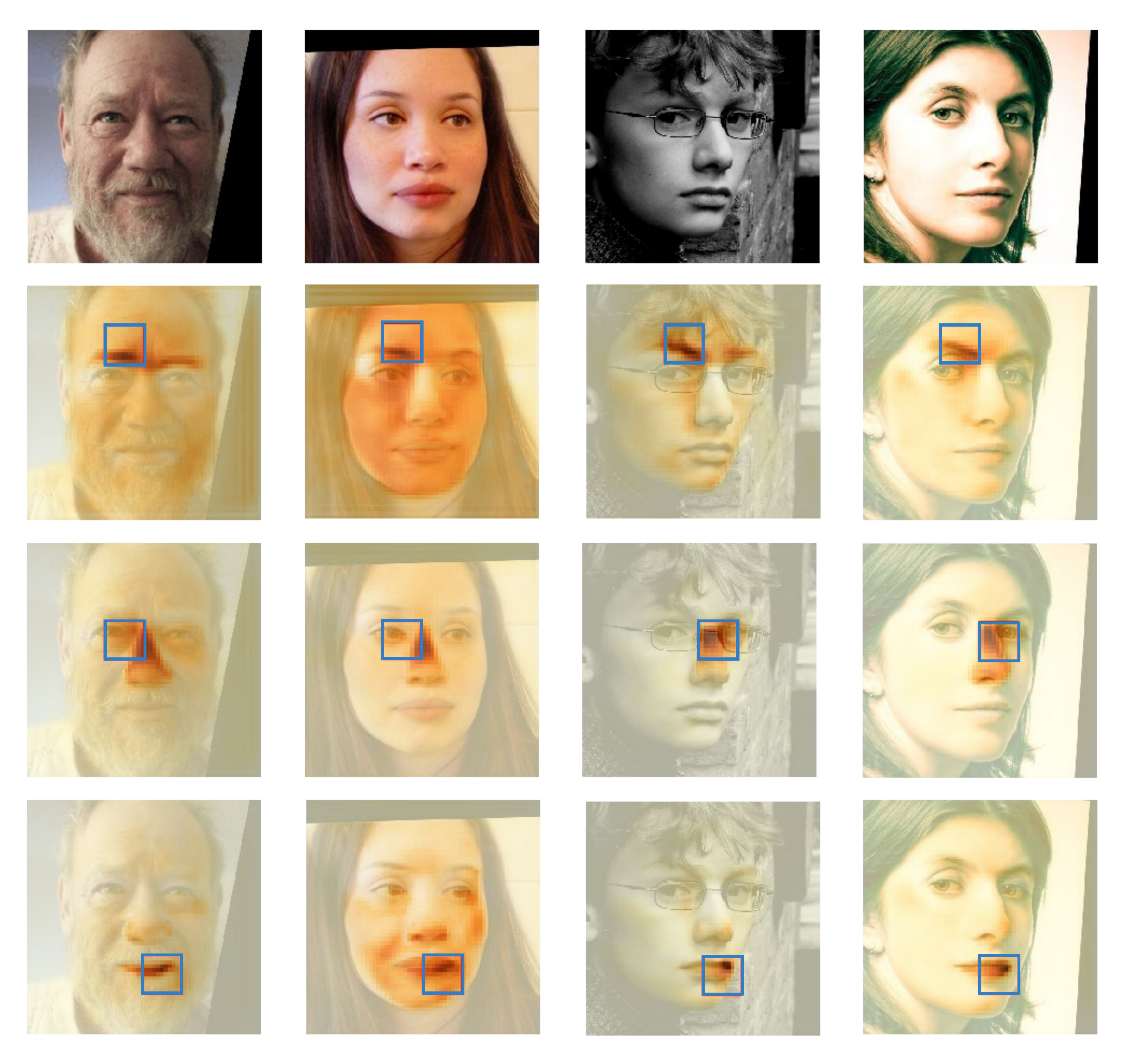}
    \caption{\textbf{Visualization of graph projection via response maps.} The first row shows the input image, and the rest visualize response maps with respect to the vertex marked in a blue rectangle. Darker color indicates higher response.  }
    \label{fig:mat_vis}
\end{figure}

Further, we visualize the graph projection for intuitive interpretation. 
As in Fig.~\ref{fig:mat_vis}, given each input image (first row), we visualize the weight of each pixel that contributes to a vertex marked in a blue rectangle in the other rows, which we refer to as the response map. 
Darker color indicates higher response. 
We observe that the response areas are consistent with the vertex, which validates that our graph projection maps pixels in the same semantic component to the same vertex. 


\newpage

\section{Conclusion}

We propose a novel graph representation learning paradigm of edge aware graph reasoning for face parsing, which captures region-wise relations to model long-range contextual information. Edge cues are exploited in order to project significant pixels onto graph vertices on a higher semantic level. 
We then learn the relation between vertices (regions) and reason over all vertices to characterize the semantic information. 
Experimental results demonstrate that the proposed method sets the new state-of-the-art with low computation complexity, which efficiently reconstructs boundary details in particular. 
In future, we will apply the paradigm of edge aware graph reasoning to more segmentation applications, such as scene parsing.

\clearpage
\bibliographystyle{splncs}
\bibliography{egbib}
\end{document}